\documentclass[journal]{IEEEtran}
\usepackage{cite}
\usepackage[T1]{fontenc}          
\usepackage{newtxtext,newtxmath}  

\usepackage{amsmath,amsfonts}
\usepackage{algorithmic}
\usepackage{graphicx}
\usepackage{subcaption}
\usepackage{caption}
\usepackage{textcomp}
\usepackage{xcolor}
\usepackage{flushend}
\usepackage{url}
\usepackage{pslatex}

\newcommand{\blue}{\color{blue}}

\begin{document}

\title{{\small \blue This paper has been accepted for publication in the proceedings of OCEANS'2025, Great Lakes}\\ \vspace{-0pt}
\Huge {SMART-OC: 
A Real-time Time-risk Optimal Replanning Algorithm
for Dynamic Obstacles and Spatio-temporally Varying Currents}\\ \vspace{10pt}
}

\author{\begin{tabular}{cccccccccc}
{Reema Raval$^\dag$} & {Shalabh Gupta$^\dag$$^\star$}
\end{tabular}\vspace{-3pt}
\thanks {$^\dag$Dept. of Electrical and Computer Engineering, University of Connecticut, Storrs, CT 06269, USA.}
\thanks {$^\star$Corresponding author (email: shalabh.gupta@uconn.edu)} \vspace{-12pt}
\thanks{Copyright 2025 IEEE.  Personal use of this material is permitted.  Permission from IEEE must be obtained for all other uses, in any current or future media, including reprinting/republishing this material for advertising or promotional purposes, creating new collective works, for resale or redistribution to servers or lists, or reuse of any copyrighted component of this work in other works.}
}

\maketitle

\thispagestyle{empty}
\section{Abstract}
Typical marine environments are highly complex with spatio-temporally varying currents and dynamic obstacles, presenting significant challenges to Unmanned Surface Vehicles (USVs) for safe and efficient navigation. Thus, the USVs need to continuously adapt their paths with real-time information to avoid collisions and follow the path of least resistance to the goal via exploiting ocean currents. In this regard, we introduce a novel algorithm, called \textbf{S}elf-\textbf{M}orphing \textbf{A}daptive \textbf{R}eplanning \textbf{T}ree for dynamic \textbf{O}bstacles and \textbf{C}urrents (SMART-OC), that facilitates real-time time-risk optimal replanning in dynamic environments. SMART-OC integrates the obstacle risks along a path with the time cost to reach the goal to find the time-risk optimal path. The effectiveness of SMART-OC is validated by simulation experiments, which demonstrate that the USV performs fast replannings to avoid dynamic obstacles and exploit ocean currents to successfully reach the goal.

\section{Introduction}
\begin{figure}
         \centering
         \includegraphics[width=0.95\linewidth]{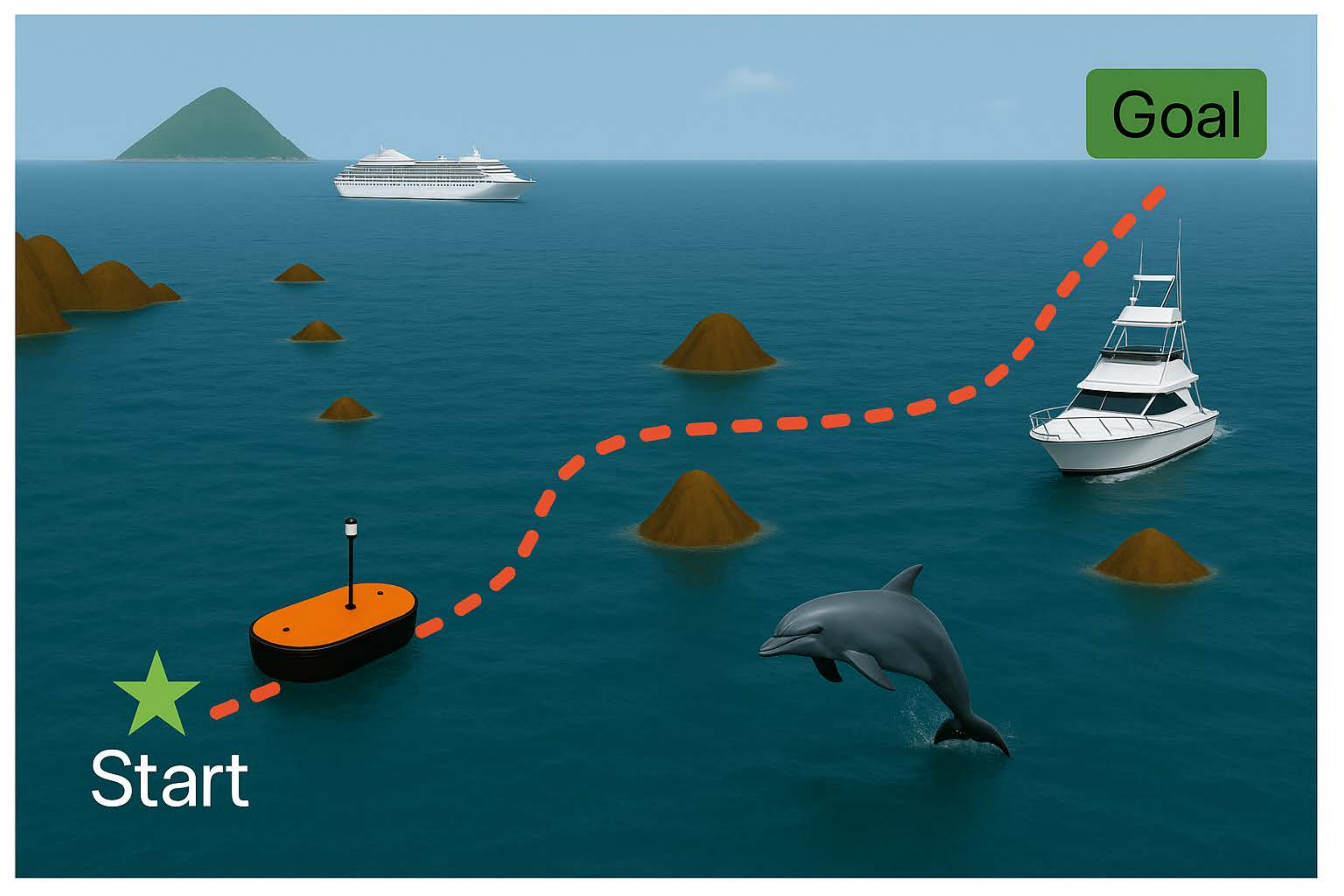}
         \caption{Complex ocean environment for USV operation.}
         \label{ComplexEnv.pdf:enter-label}\vspace{-6pt}
\end{figure}

Unmanned Surface Vehicles (USVs) have become increasingly vital in various maritime applications~\cite{Yuh2011_Applications_marinerobot} including marine life exploration~\cite{Katzschmann2018}, seafloor mapping~\cite{shen2022ct,palomeras2018,song2018,Negahdaripour2003}), data collection\cite{Somers2016}, oil spill cleaning~\cite{SGH13,SongQiu2014,Kumar2020}; human-robot collaborative tasking~\cite{YangWilson2023}, target tracking~\cite{Shojaei2017,Hare2020} and demining~\cite{mukherjee2011symbolic,Acar2003}. Modern USVs are designed with autonomous functionalities and adaptability.
However, safe and reliable navigation of USVs is challenging in complex marine environments, shown in Fig.~\ref{ComplexEnv.pdf:enter-label}, characterized by static obstacles (e.g., islands, rocks, buoys and anchored vessels),
dynamic obstacles (e.g., boats, USVs,
floating ice, and  marine wildlife) and ocean currents.

Traditional planning algorithms including graph-based search methods, probabilistic roadmaps, and artificial potential fields,
are typically optimized for static scenarios and lack responsiveness to dynamic obstacles and currents, leading to inefficient paths and increased collision risks.  This emphasizes the need for integrating dynamic obstacles and real-time oceanographic data, especially ocean currents, into path planning frameworks to enhance the navigational performance of USVs.

In this regard, this paper introduces a novel algorithm called, \textbf{S}elf-\textbf{M}orphing \textbf{A}daptive \textbf{R}eplanning \textbf{T}ree for dynamic \textbf{O}bstacles and \textbf{C}urrents (SMART-OC). SMART-OC performs real-time time-risk optimal~\cite{Songgupta2019} replanning in response to collision risks and changing ocean currents. Thus, SMART-OC provides adaptability, operational safety, and navigation efficiency for USVs operating in complex marine environments.

\begin{figure*}[t]
    \centering
    \begin{subfigure}{0.32\textwidth}
        \includegraphics[width=\linewidth]{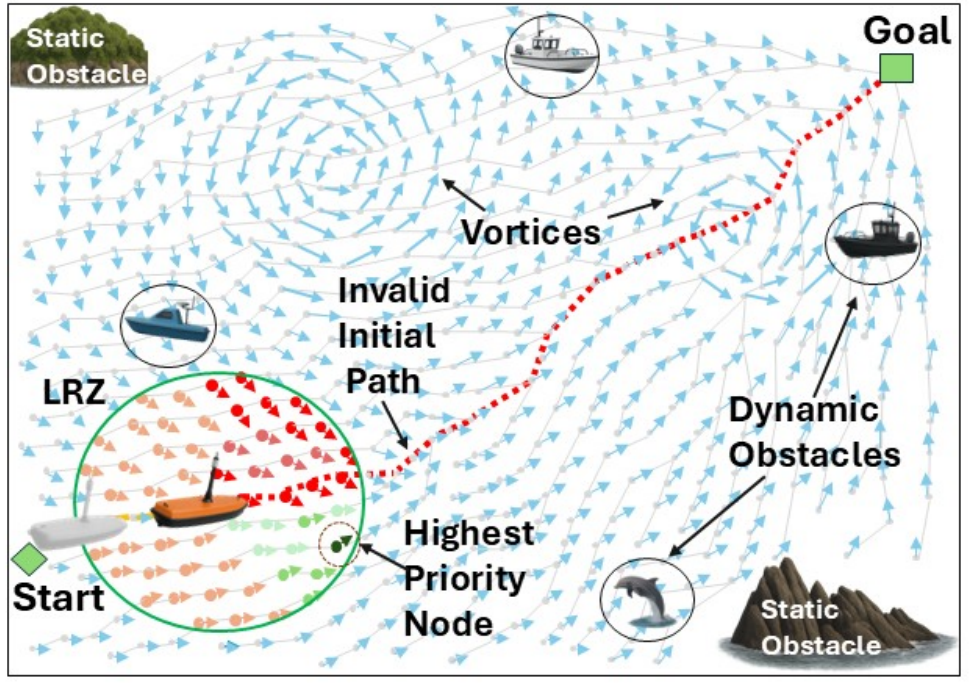}
        \caption{Initial path becomes suboptimal due to more favourable currents.}\label{fig:replan_a}
    \end{subfigure}
    \hfill
    \begin{subfigure}{0.32\textwidth}
        \includegraphics[width=\linewidth]{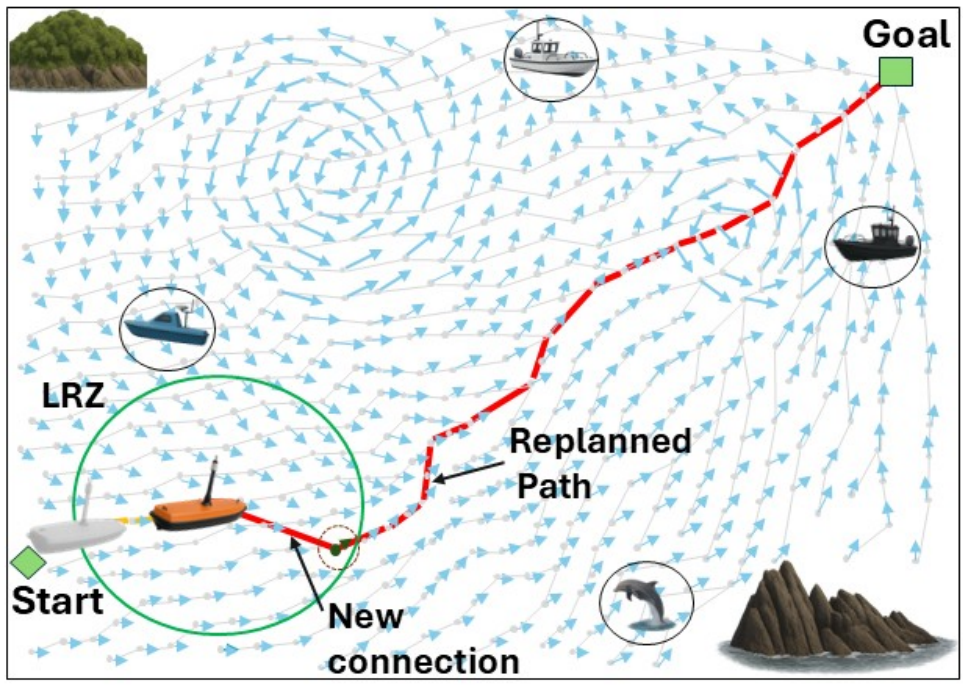}
        \caption{Replanned path using the least time-risk cost node within LRZ.}
        \label{fig:replan_b}
    \end{subfigure}
    \hfill
    \begin{subfigure}{0.32\textwidth}
        \includegraphics[width=\linewidth]{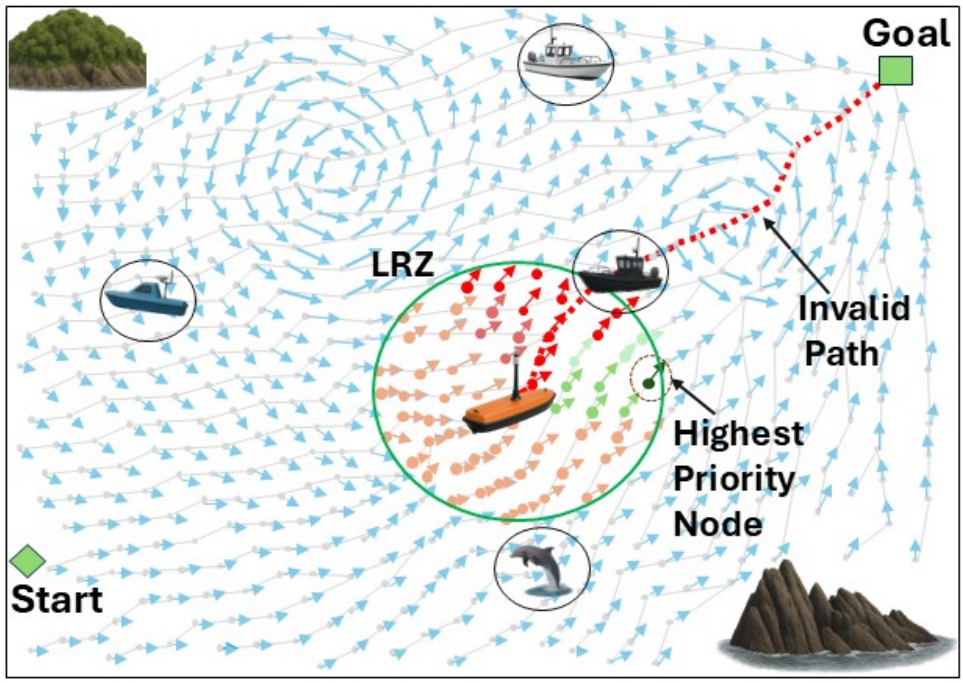}
        \caption{Dynamic obstacle increases collision risk which triggers replanning.}
        \label{fig:replan_c}
    \end{subfigure}

    \vspace{0.5em}

    \begin{subfigure}{0.32\textwidth}
        \includegraphics[width=\linewidth]{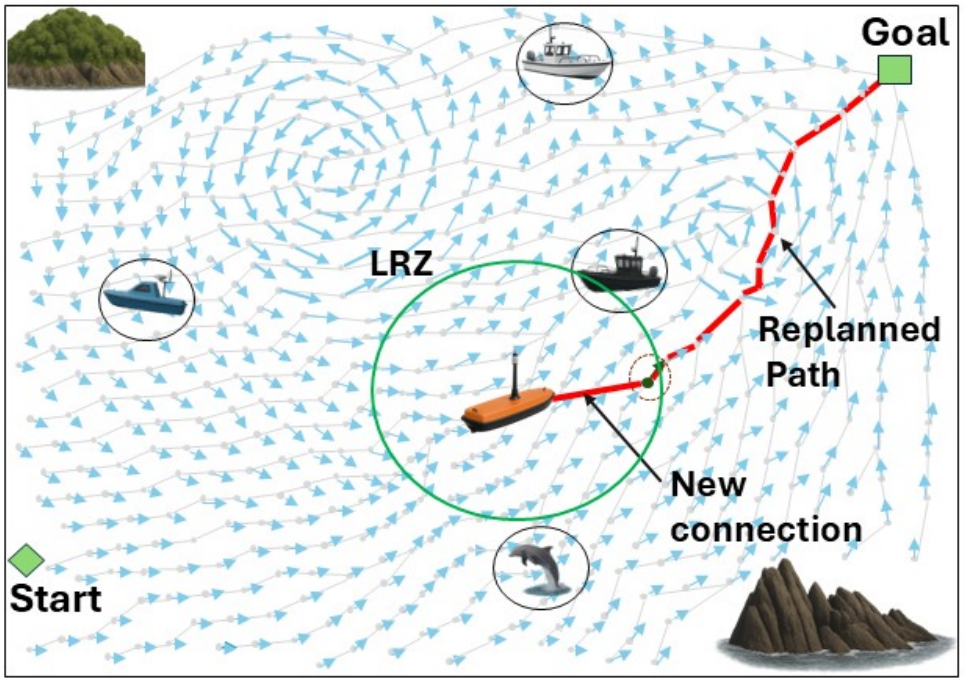}
        \caption{Replanned path avoiding obstacle by selecting the highest priority node within LRZ.}
        \label{fig:replan_d}
    \end{subfigure}
    \hfill
    \begin{subfigure}{0.32\textwidth}
        \includegraphics[width=\linewidth]{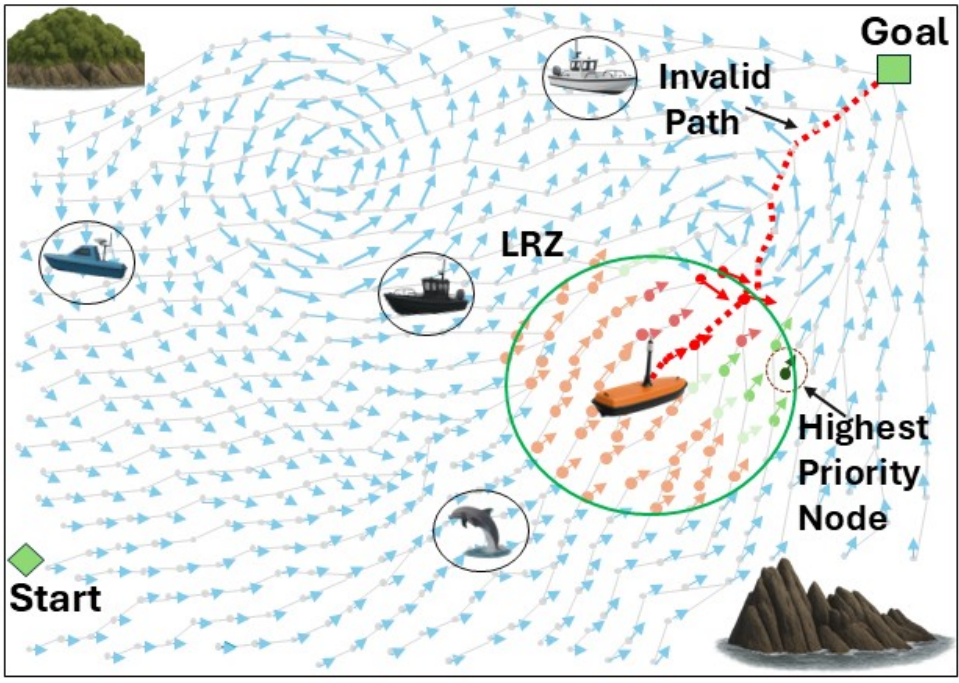}
        \caption{Path becomes invalid again due to adverse currents which triggers replanning.}
        \label{fig:replan_e}
    \end{subfigure}
    \hfill
    \begin{subfigure}{0.32\textwidth}
        \includegraphics[width=\linewidth]{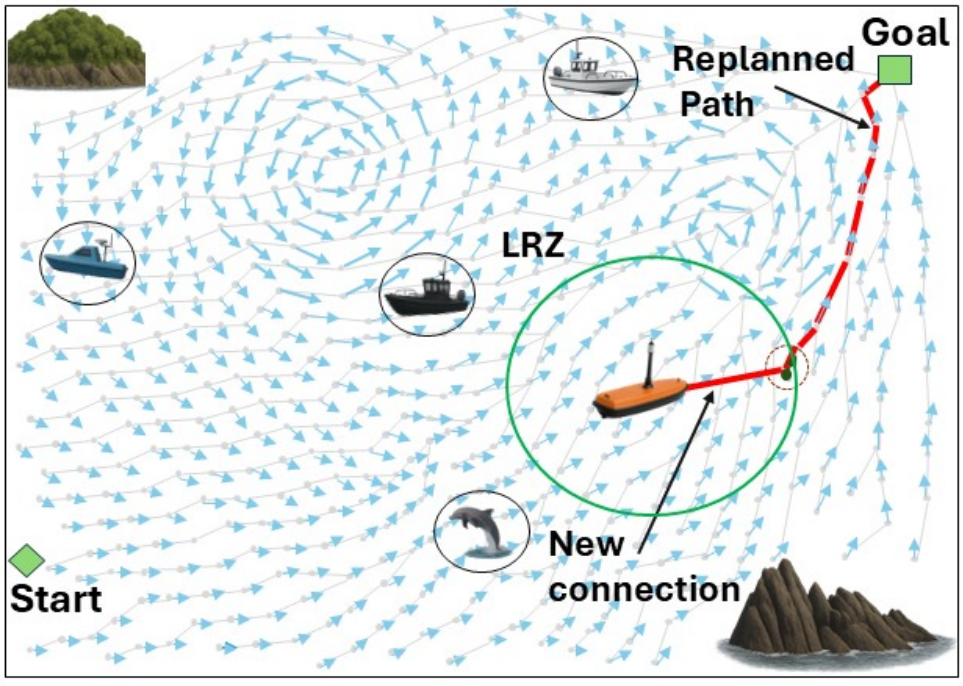}
        \caption{Final replanned path is computed based on highest priority node selection.}
        \label{fig:replan_f}
    \end{subfigure}
    \caption{SMART-OC continuous replanning under dynamic obstacles and spatio-temporally varying ocean currents.}
    \label{fig:smart_oc}\vspace{-6pt}
\end{figure*}

\vspace{-0pt}
\section{Related Work}
Recent research on USV path planning 
addresses some of the challenges posed by dynamic ocean environments and obstacles. Duan et al.\ \cite{Duan2024} used weighted dynamic programming for time‐optimal, obstacle aware routes without modeling full reactivity to dynamic obstacles and ocean current.
Zhang et al.~\cite{Zhang2023SVFRRT} applied a stream-function heuristic in RRT* for energy-efficient current exploitation but still require full rewiring and do not explicitly handle dynamic obstacles. Zhang et al.~\cite{Zhang2016CoastalAStar}  augmented A*
with vortex dynamics model and B$'$ezier smoothing without onboard real-time current update or a reactive framework.
Park et al.\ \cite{Park2023ASV} fused risk estimation and Dubins-curve avoidance while treating currents as mere disturbances.
Peng et al.\ \cite{Peng2024MultiTask3D} combined Ant Colony, PSO, and RRT for 3D missions at high algorithmic cost. Mittal and Gupta \cite{MittalGupta2019} developed an optimal-control hybrid A* planner under currents with offline preprocessing and another real-time planner for Dubins vehicles~\cite{mittal2020rapid}.
Wilson et al.\ \cite{WG2025} propose variable‐speed model for non-holonomic vehicles for time-risk optimization. Gupta et al.~\cite{GRP09} proposed an adaptive method based on Ising model for underwater target hunting.

Recently, a real-time method called SMART \cite{Shen2023SMART}, was developed to facilitate reactive replanning for dynamic obstacles by performing local risk-based tree‐pruning, and quick repairing at identified hotspots; however, SMART does not consider ocean currents, which are critical for USV navigation in marine environments. Chen et al.~\cite{12} proposed a hybrid path planning algorithm combining global A* search with the Dynamic Window Approach (DWA) for local obstacle avoidance. Nonetheless, their method incurs substantial computational overhead due to repeated global path recalculations and does not explicitly incorporate real-time updates of dynamic ocean currents.

Fuller and Thein~\cite{13} developed a kinodynamic Rapidly-exploring Random Tree (RRT) method emphasizing dynamically feasible trajectory generation for Autonomous Surface Vehicles (ASVs), yet their approach lacks explicit consideration of real-time adaptability to dynamic ocean conditions and obstacle movements. Chiang and Tapia~\cite{14} introduced a COLREGS-compliant RRT-based planner (COLREG-RRT), which accounts for maritime collision regulations and dynamic obstacle avoidance; however, their implementation omits integration with realistic ocean current models and real-time environmental updates. To et al.~\cite{15} provided streamline-based distance and steering heuristics within the RRT* algorithm, facilitating efficient path planning in strong flow fields, but their method is predominantly designed for steady (time-independent) flow fields only  and does not directly handle dynamic ocean currents or obstacle variability.

For real-time applications, ocean currents can be estimated through a combination of onboard and external sensing sources.
Onboard sensors include Acoustic Doppler Current Profilers (ADCPs), Inertial Navigation Systems (INS), and GPS-Derived Velocity Matching. \cite{ferreira2018multi}.
Additionally, environmental data from oceanographic forecast models
(e.g., ROMS or Copernicus Marine Service)
can be fused for predictive planning.
Sensors such as LiDAR, marine radar, and stereo vision cameras, have also been widely used for detecting obstacle hazards in marine environments
\cite{singh2021lidar}.
\cite{ song2016vision}.

While the above contributions have addressed either ocean currents or obstacle hazards, there remains an opportunity to develop a unified, computational efficient, and reactive framework that considers the effects of both spatio-temporally varying currents and dynamic obstacles.

\vspace{-0pt}
\section{SMART-OC}
To address the above limitations, we propose SMART-OC, which considers both spatio-temporally varying currents and dynamic obstacles for time-risk optimal replanning. It proactively triggers adaptive replanning upon detecting collision threats from obstacles or significant deviations in ocean currents. Then, it utilizes a a resilient node selection strategy based on a time-risk cost to compute alternate paths in real-time.
The sensing strategies can be integrated into SMART-OC deployments to enhance its effectiveness under real-world ocean conditions.
SMART-OC operates using the following steps to enable fast, safe, and current-aware replanning:

\begin{itemize}
\vspace{6pt}
\item [1.] \textbf{Initial path generation:} First, a goal-rooted RRT* tree is constructed using static obstacles data, producing a baseline path $\sigma_0$ from start $X_s$ to goal $X_g$. This distance-based optimal path ignores currents and dynamic obstacles.


\vspace{6pt}
\item [2.] \textbf{Continuous environment monitoring:} While moving on $\sigma_0$, the USV continuously monitors the ocean currents and the proximity to obstacles within a Local Reaction Zone (LRZ). Then, it computes the current deviations and collision-risks, if they exist replanning is triggered.

\vspace{6pt}
\item [3.] \textbf{Time-risk optimal replanning:} If replanning is triggered, then the time-risk cost of each node $N$ in LRZ is calculated as follows
 \begin{equation}
 \text{Cost}(N) = \frac{f(N)}{1 - r(N)},
\end{equation}
where $f(N)$ is the cumulative time-cost function consisting of cost-to-come (C(R, N)) from USV to node $N$ and cost-to-go (g(N)) from $N$ to the goal on the tree. It is given as:
\begin{equation}
f(N) = C(R, N) + g(N).
\end{equation}
The time costs are computed based on the effective velocity of the USV considering its speed and heading angle in the ocean current field. The obstacle-risk function r(n) is given as follows:
\begin{equation}
r(n) =
\begin{cases}
1, & d \leq d_{\min}, \\
\exp\left( \frac{d - d_{\min}}{d - d_{\max}} \right), & d_{\min} < d < d_{\max}, \\
0, & d \geq d_{\max},
\end{cases}
\end{equation}
where  \(d\) is the Euclidean distance between the USV and the obstacle, \(d_{\min}\) is the inner safety threshold  and \(d_{\max}\) is the outer safety threshold. Using this risk function the average path risk is calculated.
\end{itemize}

Once the above costs afre computed, the node with the minimum overall time-risk path cost is selected and reconnected to the USV and a new path is obtained from the RRT* tree.

Fig.~\ref{fig:smart_oc} shows an illustation of the workings of SMART-OC. Fig.~\ref{fig:replan_a} shows the initial RRT* path considering only static obstacles. As the USV moves, it encounters
varying ocean currents, which triggers replanning. SMART-
OC calculates the time-risk cost values for nodes within LRZ
(color coded) and obtains a new path, shown in Fig.~\ref{fig:replan_b}.
Similarly, as USV progresses, it encounters a dynamic obstacle, which triggers replanning in Fig.~\ref{fig:replan_c}. Then, it generates a new path
in Fig.~\ref{fig:replan_d}. Another replanning happens due to a vortex in Fig.~\ref{fig:replan_e} and it generates a new path in  Fig.~\ref{fig:replan_f} bef ore reaching the goal.

\vspace{3pt}
\section{Results}

\begin{figure*}[t]
  \centering
  \begin{subfigure}[t]{0.325\textwidth}
    \centering
    \includegraphics[width=\linewidth]{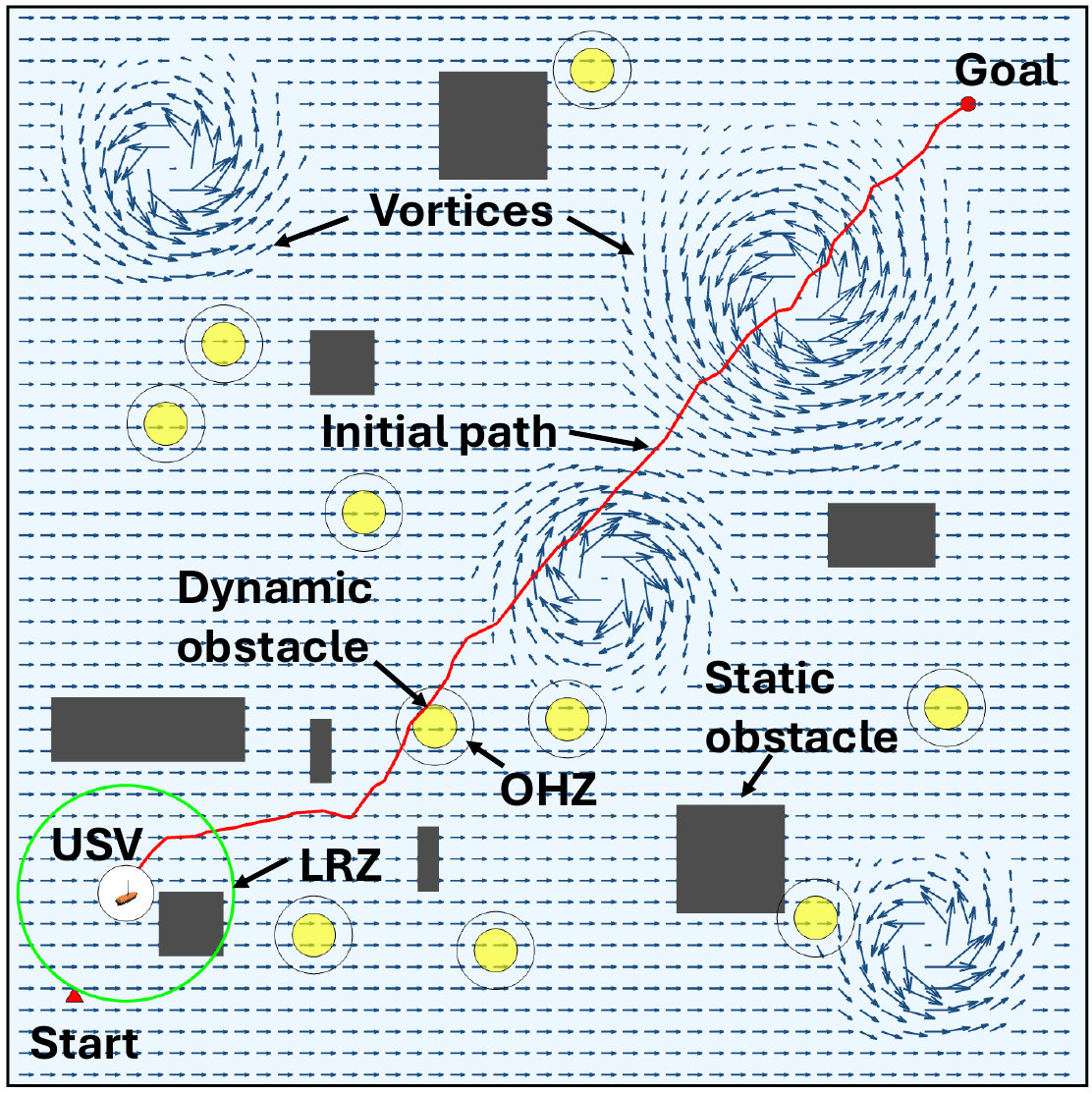}
    \caption{USV starts to move on the initial RRT* path considering only static obstacles.}
    \label{fig:scenario:a}
  \end{subfigure}\hfill
  \begin{subfigure}[t]{0.325\textwidth}
    \centering
    \includegraphics[width=\linewidth]{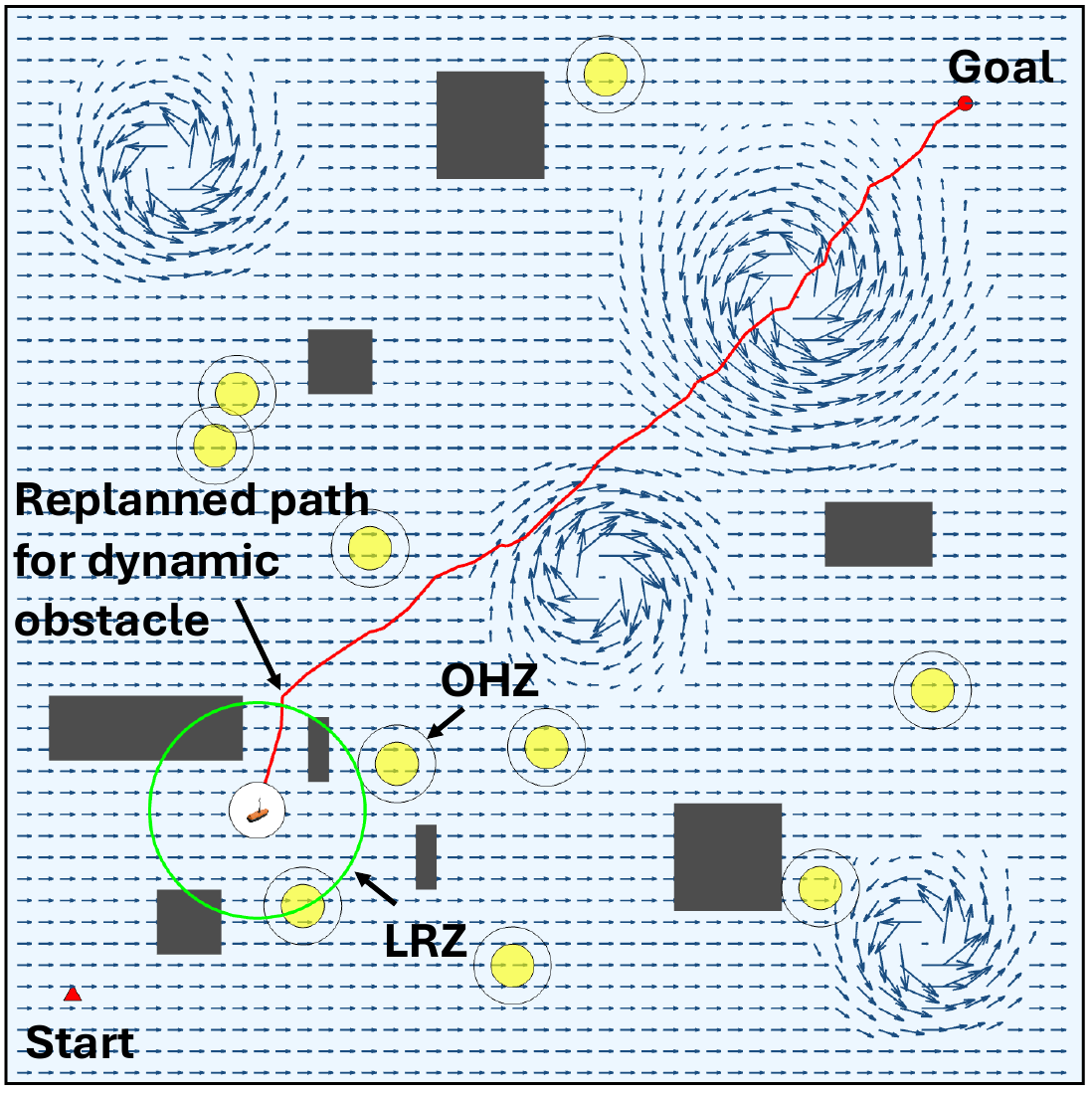}
    \caption{Replanning triggered due to a dynamic obstacle and a new path is planned.}
    \label{fig:scenario:b}
  \end{subfigure}\hfill
  \begin{subfigure}[t]{0.325\textwidth}
    \centering
    \includegraphics[width=\linewidth]{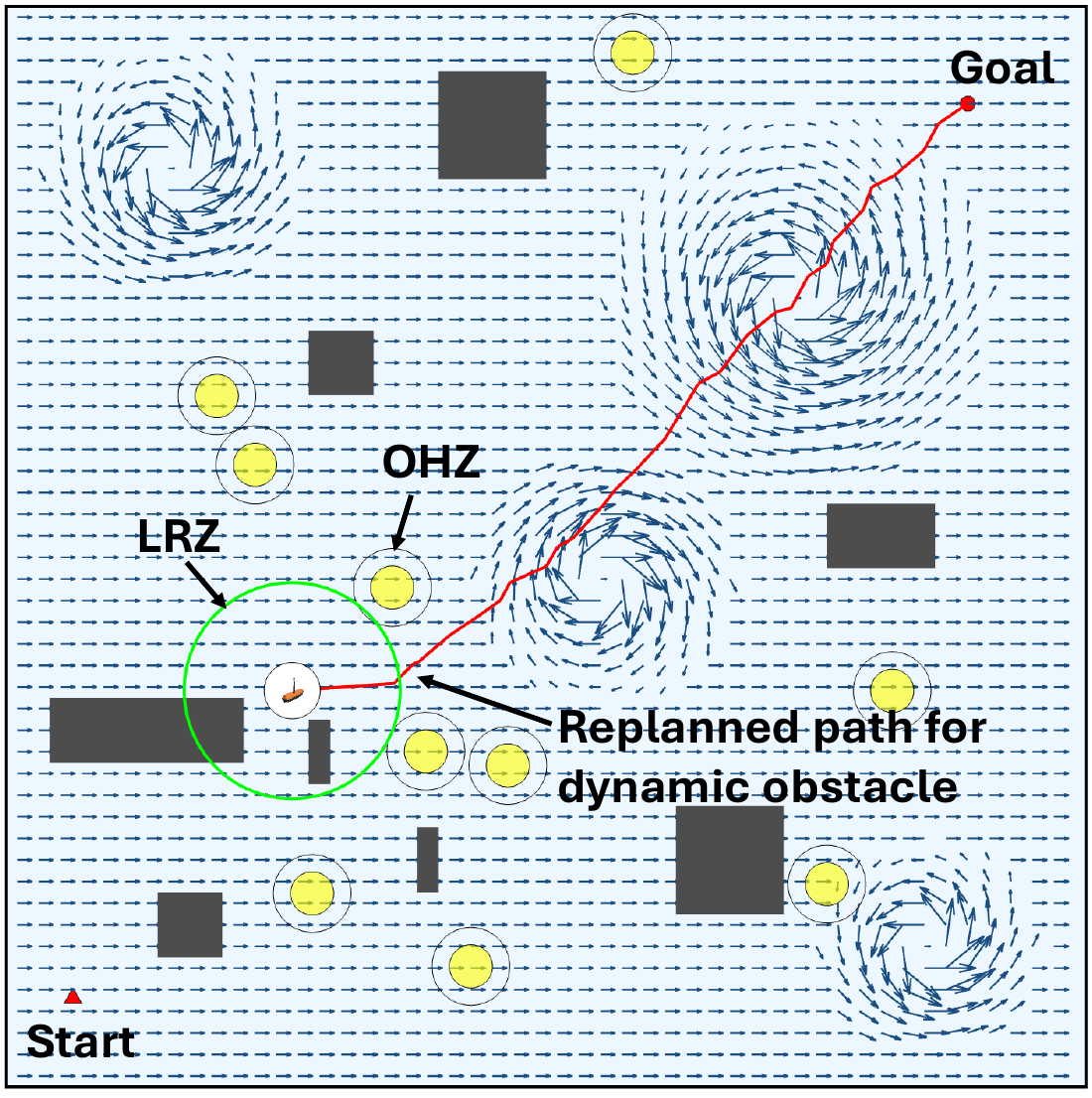}
    \caption{Replanning triggered due to another dynamic obstacle and a new path is planned.}
    \label{fig:scenario:c}
  \end{subfigure}

  \vspace{0.5em}

  \begin{subfigure}[t]{0.325\textwidth}
    \centering
    \includegraphics[width=\linewidth]{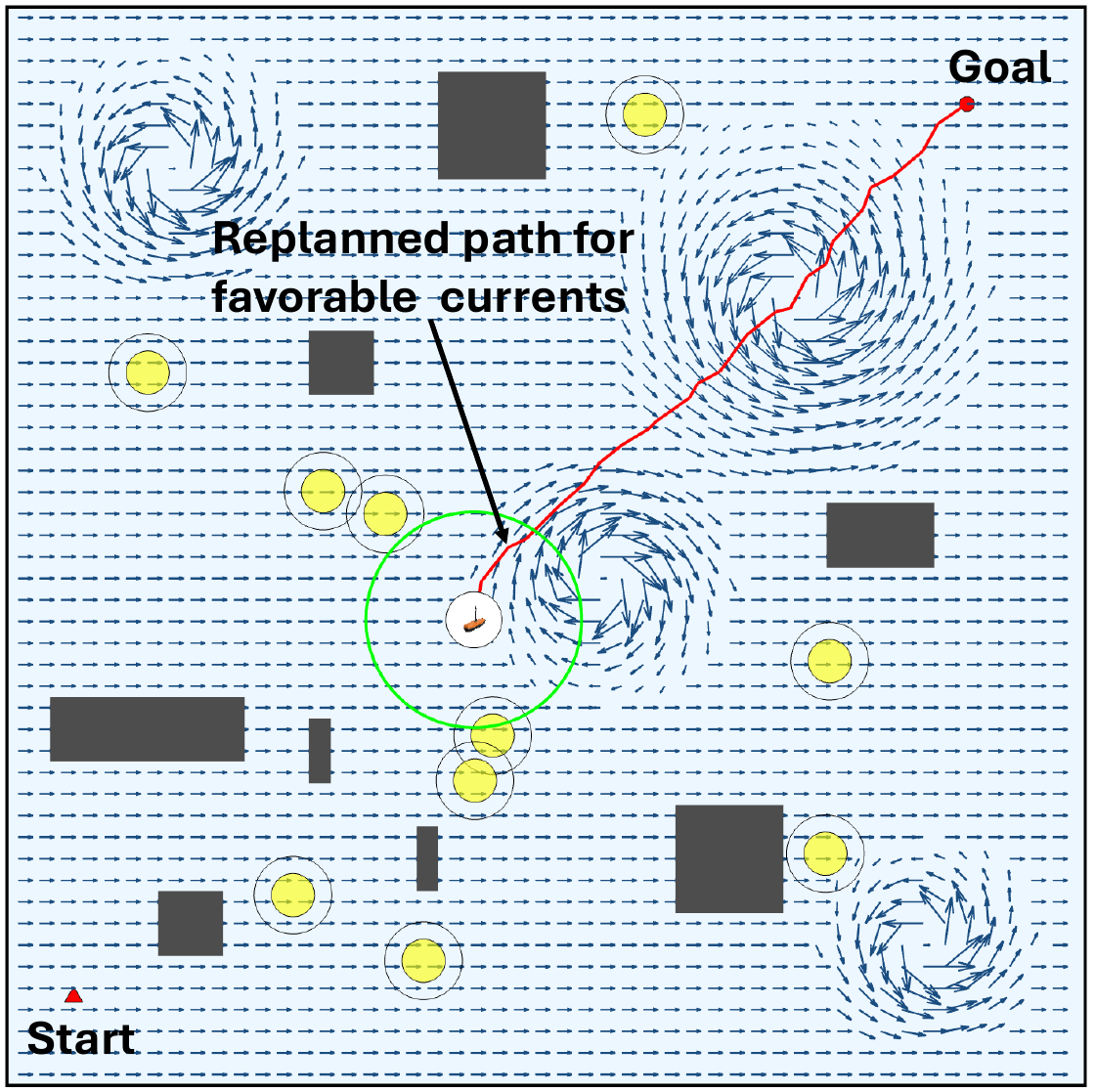}
    \caption{Replanning triggered due to a vortex and a new path is planned for favorable currents.}
    \label{fig:scenario:d}
  \end{subfigure}\hfill
  \begin{subfigure}[t]{0.325\textwidth}
    \centering
    \includegraphics[width=\linewidth]{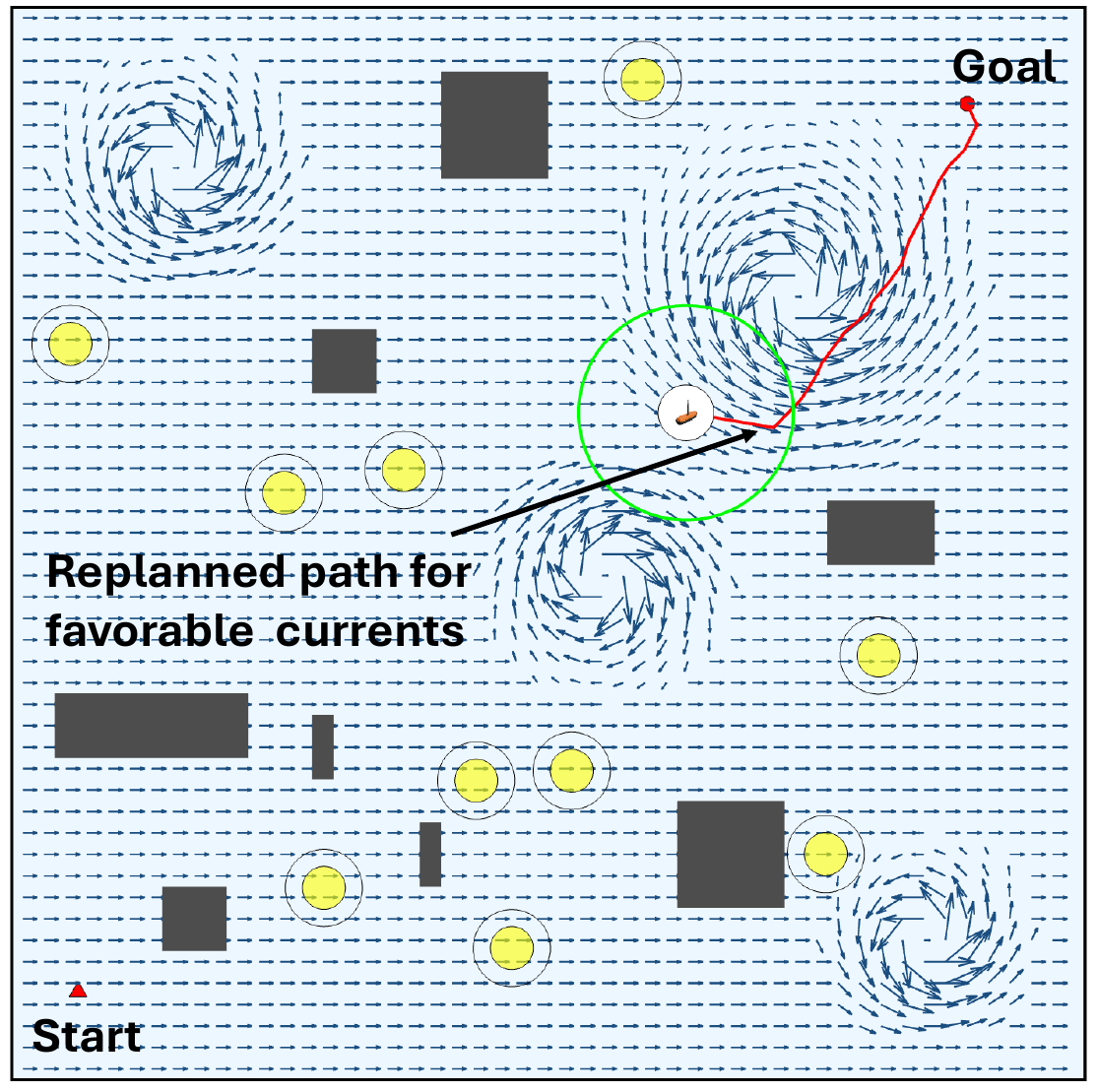}
    \caption{Replanning triggered due to a vortex and a new path is planned for favorable currents.}
    \label{fig:scenario:e}
  \end{subfigure}\hfill
  \begin{subfigure}[t]{0.325\textwidth}
    \centering
    \includegraphics[width=\linewidth]{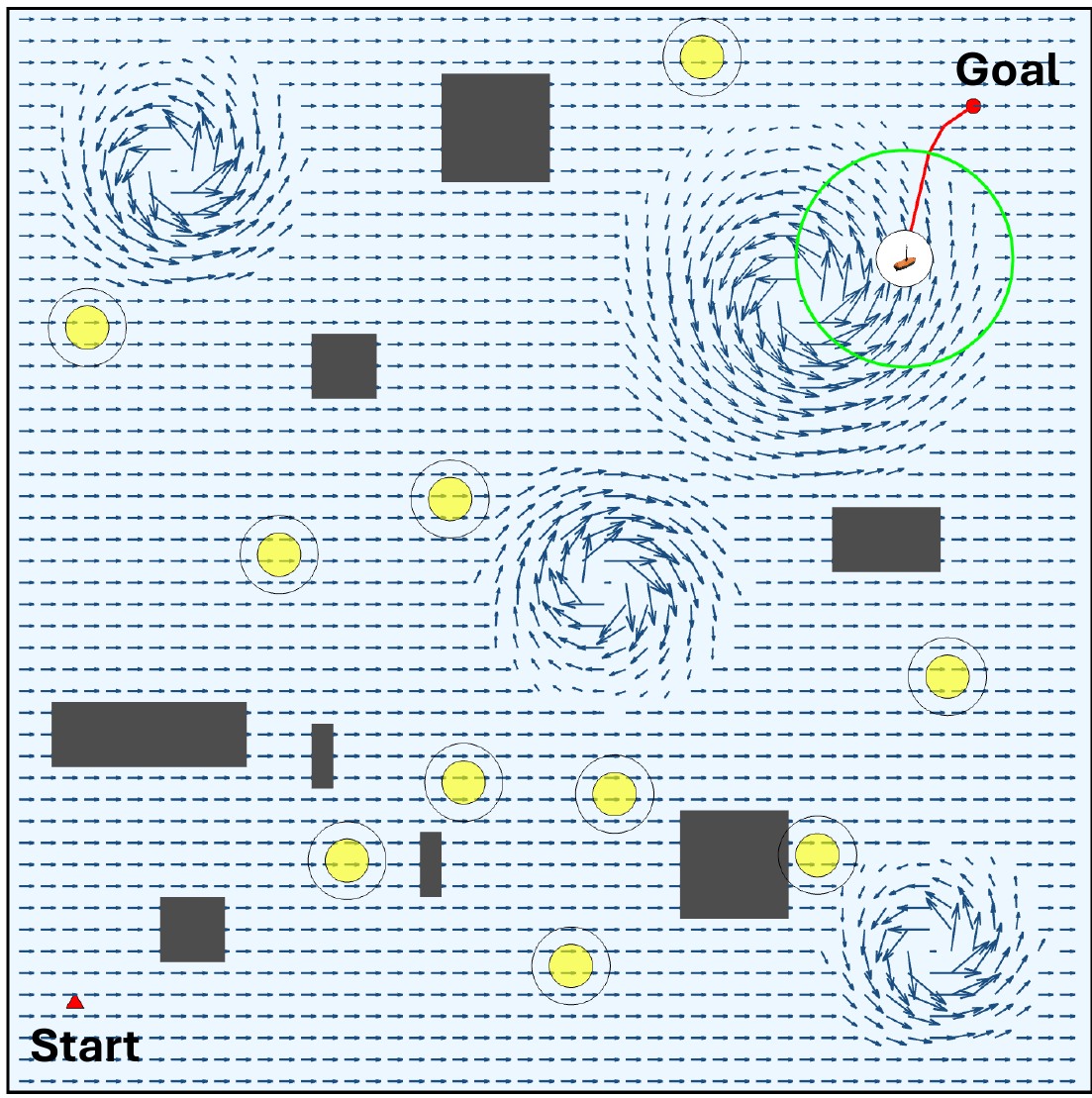}
    \caption{USV reaches the goal.}
    \label{fig:scenario:f}
  \end{subfigure}

  \caption{SMART-OC execution on a simulation platform showing several replannings for time-risk optimization of the navigation path due to dynamic obstacles and changing currents.}
  \label{fig:scenario}
\end{figure*}

This section presents the SMART-OC validation results by simulation experiments. All experiments were conducted on an Apple MacBook Pro running macOS 14 “Sonoma.”
The machine is equipped with an Apple M3 Pro chip (12-core CPU, 18-core GPU) and 16 GB of unified memory.  SMART-OC algorithm was implemented in C++ and all timing measurements (e.g. replanning time) reflect CPU execution. Visualization of results was performed in MATLAB. Table~\ref{tab:simulation_parameters} summarizes the key parameters used in the simulation environment for evaluating the SMART-OC algorithm.

Fig. {\ref{fig:scenario}  presents snapshots of simulation results of path-planning scenarios that demonstrate the effectiveness of our SMART-OC algorithm in dynamically evolving ocean environments characterized by static obstacles, moving obstacles, and spatio-temporally varying ocean currents.

Fig. {\ref{fig:scenario:a}} shows the initial environment setup consisting of the USV's start and goal locations, static obstacles (e.g., rocks), dynamic obstacles (e.g., other USVs, marine life, boats), and the background ocean current field with multiple vortices.  The grey rectangles and squares represent different static obstacles. The yellow circular structures are dynamic obstacles with Obstacle Hazard Zones (OHZs) around them. The green circle around USV is Local Reaction Zone (LRZ). The scenario consists of four different vortices from which two vortices come across the USV path. The initial path is generated using the goal-rooted RRT* tree considering only the static obstacles. The USV has started moving on this initial path.

Figs. {\ref{fig:scenario:b} and {\ref{fig:scenario:c}}} show replanning incidents triggered due to the presence of dynamic obstacles intersecting the path in LRZ. The algorithm efficiently detects the collision risks as soon as OHZ intersects with LRZ and generates new safe paths around the intruding obstacles; thereby avoiding collision with dynamic obstacles. Figs. {\ref{fig:scenario:d} and {\ref{fig:scenario:e}}} show other replanning incidents triggered due to deviations in the ocean currents within the LRZ. These replannings generate new paths, leveraging favorable flows to boost the effective speed of the USV. This demonstrates the method’s capability to maintain safety while simultaneously optimizing the path in response to dynamic environmental conditions.

Fig. {\ref{fig:scenario:f}  depicts the final phase; that is the USV approaching to the goal, successfully avoiding obstacles and utilizing favorable current flows. The resulting trajectory is consistently smooth, safe, and adheres to the vehicle’s dynamic constraints.

These results validate that SMART-OC not only responds robustly to dynamic obstacle intrusion, but also capitalizes on the real-time ocean current information to generate energy-efficient and risk-aware paths.

\begin{table}[t]
\centering
\caption{Simulation parameters}
\label{tab:simulation_parameters}
\begin{tabular}{l|c}
\hline
\textbf{Parameter} & \textbf{Value} \\
\hline
USV Speed                   & 4 m/s \\
Dynamic Obstacle Speed      & 3 m/s \\
LRZ Radius                  & 5 m \\
Current Speed Range         & 1–8 m/s \\
Number of Dynamic Obstacles & 10    \\
Number of Static Obstacles  & 8     \\
\hline
\end{tabular}
\end{table}

\begin{figure}[t]
    \centering
    \includegraphics[width=\columnwidth]{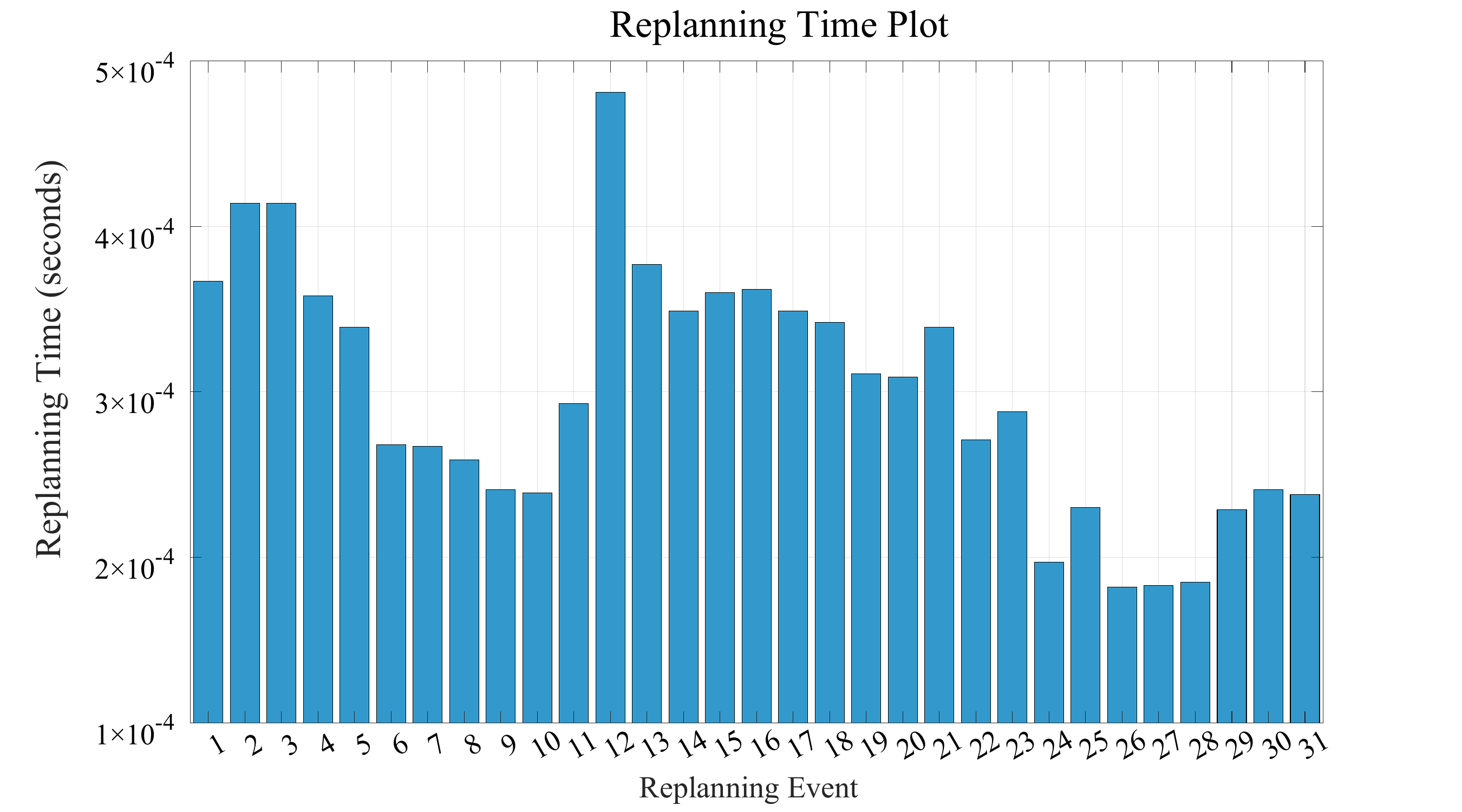}
    \caption{Replanning time for each replanning event. }
    \label{fig:replanning_time}
\end{figure}

Fig. ~\ref{fig:replanning_time} illustrates the replanning time required at each instance when the SMART-OC algorithm is triggered during navigation. The replanning incidents and replanning  times are plotted on the X and Y axes, respectively. This replanning time plot shows that the operations are completed within a very short time window, demonstrating the efficiency and real-time capability of the SMART-OC algorithm. The consistency across all 31 replanning events with average replanning time of $2.9\times10^{-4}$ seconds reflects the algorithm’s computational efficiency and real-time suitability for autonomous USV operations in dynamic ocean environments.

\vspace{-3pt}
\section{Conclusion and Future Work}
This paper presented SMART‑OC, a real-time replanning algorithm tailored for USVs navigating through uncertain and dynamic marine environments. Unlike traditional planners that rely on static maps or delayed updates, SMART‑OC adapts to both dynamic obstacles and current disturbances by actively monitoring the local surroundings and selectively restructuring its search tree for selecting minimum time-risk cost path. This allows SMART‑OC to maintain feasible, safe, and minimum time trajectories throughout the mission.

While the simulation results validate SMART‑OC’s robustness and responsiveness under challenging ocean conditions, there are several directions for further improvement. Future work will include extending SMART‑OC for collaborative multi-USV planning~\cite{song2020care} and incorporating energy-aware optimization~\cite{shen2020} to further improve mission longevity and efficiency in resource-constrained scenarios.

\vspace{-0pt}
\bibliographystyle{IEEEtran}
\bibliography{SMART-OC}

\end{document}